\pdfoutput=1
\documentclass{article}

\usepackage{arxiv}

\usepackage[utf8]{inputenc}
\usepackage[T1]{fontenc}
\usepackage{hyperref}
\usepackage{url}
\usepackage{booktabs}
\usepackage{amsmath,amssymb,amsfonts}
\usepackage{array}
\usepackage{nicefrac}
\usepackage{microtype}
\usepackage{graphicx}
\usepackage[numbers]{natbib}
\usepackage[most]{tcolorbox}
\usepackage{tikz}
\usepackage{orcidlink}
\usetikzlibrary{positioning, arrows.meta, shapes.geometric, fit, backgrounds, calc}

\definecolor{qheader}{RGB}{40,40,40}
\definecolor{qlight}{RGB}{245,245,245}

\newtcolorbox{trajectorycard}[1]{
  colback=white,
  colframe=qheader,
  coltitle=white,
  fonttitle=\small\bfseries,
  title={#1},
  boxrule=0.6pt,
  arc=1pt,
  left=6pt, right=6pt, top=4pt, bottom=4pt,
  toptitle=2pt, bottomtitle=2pt,
}

\graphicspath{{figures/}}

\title{Trustworthy Multi-Agent Systems:\\Mitigating Semantic Drift with the Argent Signaling Protocol}

\author{
  Anantha Sharma\,\orcidlink{0000-0002-9064-3362} \\
  Synechron Inc \\
  Charlotte, NC, USA
}
\renewcommand{\undertitle}{ }
\renewcommand{\headeright}{}

\date{April 16, 2026}

\renewcommand{\shorttitle}{Mitigating Semantic Drift with ASP}

\hypersetup{
  pdftitle={Trustworthy Multi-Agent Systems: Mitigating Semantic Drift with the Argent Signaling Protocol},
  pdfauthor={Anantha Sharma},
  pdfkeywords={Multi-Agent Systems, Explainable AI, Semantic Drift, Uncertainty, Grounded Generation, Large Language Models},
}

\begin{document}
\maketitle

\begin{abstract}
When multi-agent LLM systems produce bad answers, not all failures are equal: some answers are grounded in the right material but incomplete, while others are simply ungrounded and should be stopped. Current retry strategies treat both cases identically (try again and hope for the best), leaving human supervisors unable to tell whether a retry was warranted or whether the system should have halted instead.

We introduce the \emph{Argent} Signaling Protocol (ASP) \emph{argent} in the sense of silver, signaling clarity and audit-grade transparency, and distinct from Answer Set Programming. ASP is a compact machine-readable header that accompanies every AI-generated response with structured quality signals: certainty (@C), grounding (@G), stochasticity (@S), and an assumption index that classifies the evidentiary basis of each claim. These signals enable a controller to distinguish \emph{repairable} failures from \emph{containment} failures and route each case differently.

We evaluate ASP in two modes. In standalone mode, a 27-question document-grounded QA benchmark over the Array BioPharma/Ono license agreement compares baseline prompts against ASP-instrumented controller actions across three local GGUF models. On Qwen~(0.8B), ASP improves pass rate from 11.1\% to 33.3\% and mean term coverage from 36.7\% to 65.4\%; on Dobby~(8B), ASP produces 4 fail-to-pass recoveries, raising pass rate from 33.3\% to 44.4\%; on SmolLM3~(3B), ASP alternates between repair and containment per question. Aggregate improvement is meaningful (12/81 to 21/81 passes). In multi-agent mode, an ASP sidecar sits between a retrieval agent and a downstream decision agent; the sidecar blocks 100\% of ungrounded upstream outputs from reaching the downstream agent (24/27 blocked, 0 ungrounded propagations).
\end{abstract}


\section{Introduction}

As Large Language Models (LLMs) move from isolated prompting to orchestrated Multi-Agent Systems (MAS), those intermediate outputs increasingly function as operational state rather than disposable prose \citep{autogen,camel}. In \emph{document-grounded} settings, where a model must answer questions by citing a specific source document, a downstream agent may consume an upstream answer even when that answer is only partially supported or is already drifting away from the source text.

\paragraph{A concrete example}
Consider an AI system asked to answer legal questions about a pharmaceutical license agreement. The system retrieves relevant sections of the contract and asks a language model to summarize the answer with citations. In one case, the model cites the correct section (``Section 1.6, Change in Control'') and covers two of four key terms but misses the branches about mergers and asset sales. In another case, the model produces fluent legal-sounding prose that bears no relation to any retrieved section: no citations, no grounding, just confident-looking fabrication. These are fundamentally different failures. The first answer is \emph{repairable}: the right material was retrieved, and a second attempt might fill in the missing branches. The second answer requires \emph{containment}: no amount of retrying will ground an answer that was never attached to the source material. Yet most AI retry strategies treat both identically: try again and hope for a better answer.

\paragraph{Semantic drift}
We call this accumulation of untracked assumptions \emph{semantic drift}: locally plausible steps compound into globally invalid trajectories because downstream agents consume upstream assertions as settled facts. This problem is related to known calibration and uncertainty issues in neural models \citep{guo_calib,kadavath}, but multi-agent systems amplify it by recursively reusing model outputs as fresh context. The result is not just incorrect final answers; it is a loss of auditability about whether the system is repairing grounded partial answers, polishing nonsense, or silently giving up.

\paragraph{Implications for auditability}
Explainable and risk-managed deployments require traceability, interpretable failure state, and controls that are measurable rather than purely rhetorical \citep{arrieta_xai,nist_rmf}. Yet common retry strategies still rely on natural-language feedback alone. A human auditor can often tell that the system retried, but not whether the retry was warranted, what failure class triggered it, or whether halting would have been the safer choice. Standardized reporting frameworks such as Datasheets for Datasets \citep{datasheets} and FactSheets \citep{factsheets} have established the principle that AI artifacts should carry structured metadata about their provenance and limitations. ASP extends that principle from datasets and services to individual agent-to-agent messages.

\paragraph{Our approach}
We address this gap with the Argent Signaling Protocol (ASP), a lightweight protocol that attaches structured quality signals to every AI-generated response. ASP builds on the Argent Framework \citep{sharma_argent}, which introduced governance-first primitives for composing, orchestrating, and governing enterprise AI agents. Where the Argent Framework defines how agents are built and bounded, ASP defines how they communicate their confidence, grounding, and assumptions. This enables the governing controller to make informed routing decisions rather than relying on opaque retry loops.

This paper goes into the following areas:
\begin{itemize}
    \item Define the Argent Signaling Protocol: a compact, auditable header that externalizes certainty, grounding, stochasticity, and an assumption index for each AI response.
    \item Describe a sidecar deployment model in which ASP operates as an interception layer between agents, enforcing deterministic quality gates without modifying the agents themselves.
    \item Demonstrate ASP in standalone mode (single agent with controller loop), producing 10 fail-to-pass recoveries across three models (12/81 to 21/81 aggregate passes).
    \item Demonstrate ASP in multi-agent mode (two-agent pipeline with sidecar gating), showing that the sidecar blocks 100\% of ungrounded upstream outputs from reaching a downstream decision agent.
    \item Show that the same controller adapts its role to the model it governs (repair on competent models, per-question triage between repair and containment on a mixed model).
\end{itemize}

\paragraph{Scope}
The evaluation covers both operating modes. The standalone controller benchmark (Section~\ref{sec:controller-results}) tests ASP within a single agent's retry loop over a document-grounded QA task. The pipeline benchmark (Section~\ref{sec:pipeline-results}) tests ASP as a sidecar between two agents, demonstrating that the protocol prevents ungrounded upstream outputs from propagating to a downstream decision agent.

\section{Related Work}

\paragraph{Agent frameworks}
CAMEL \citep{camel} and AutoGen \citep{autogen} show that multi-agent decompositions can attack problems that a single LLM call cannot. Both exchange free-form language or task-specific tool calls; neither carries a standardized uncertainty or provenance contract. When Agent~A tells Agent~B ``the contract defines Change in Control in Section~1.6,'' Agent~B cannot tell whether Agent~A was quoting retrieved evidence or generating from memory, and has no structured way to verify the claim before acting on it. The Argent Framework \citep{sharma_argent} addresses agent composition and budget-bounded execution at the orchestration level; ASP occupies the protocol layer between the agent and its governor.

\paragraph{Self-correction}
Self-Refine \citep{selfrefine} and Reflexion \citep{reflexion} demonstrate that feedback-driven retries can improve answers. The limitation in high-stakes settings is that the feedback stays rhetorical: after three retries, an auditor sees that the system tried three times but cannot tell from the log whether any attempt was grounded, whether confidence rose or fell across attempts, or whether halting earlier would have been safer. ASP makes those signals explicit and machine-readable, so every retry carries its own quality report card.

\paragraph{Calibration and uncertainty}
Modern neural networks are often miscalibrated \citep{guo_calib}; LLMs sometimes know what they know under the right elicitation protocol \citep{kadavath}; and verbalized confidence can diverge from true correctness when asked for in natural language alone \citep{kuhn_unc,xiong_unc}. ASP does not claim access to true posterior confidence. It operates on auditable proxies of grounding and instability and asks whether those proxies are sufficient, in practice, to route between repair and containment. The protocol's contribution is to push these signals from research benchmarks into the communication boundary between agents, where they can drive routing rather than only post-hoc analysis.

\paragraph{Documentation and reporting standards}
Datasheets for Datasets \citep{datasheets} standardize documentation for training data; FactSheets \citep{factsheets} do the same for AI services. ASP extends that practice to the inter-agent communication layer: each message carries its own provenance, confidence, and assumption metadata, so the entire multi-agent conversation is auditable as a chain of documented decisions rather than as opaque text.

\paragraph{Where ASP sits in the design space}
Table~\ref{tab:related-comparison} compares ASP to the closest related systems along four axes: whether the inter-agent protocol carries calibrated uncertainty, whether it carries a provenance/assumption channel, whether the orchestrator monitors drift between turns, and whether the system can circuit-break with an auditable record of why. To our knowledge ASP is the first agent protocol that exposes all four at the message boundary.

\begin{table}[t]
\centering
\caption{ASP versus closely related multi-agent and self-correction systems. $\checkmark$ = present at the protocol or orchestration level; $\times$ = absent; partial = available as natural-language critique without machine-readable fields}
\label{tab:related-comparison}
\small
\begin{tabular}{lcccc}
\toprule
\textbf{System} & \textbf{Calibrated} & \textbf{Provenance /} & \textbf{Drift} & \textbf{Auditable} \\
                & \textbf{uncertainty} & \textbf{assumption channel} & \textbf{monitor} & \textbf{circuit-break} \\
\midrule
AutoGen \citep{autogen}        & $\times$ & $\times$ & $\times$ & $\times$ \\
CAMEL \citep{camel}            & $\times$ & $\times$ & $\times$ & $\times$ \\
Self-Refine \citep{selfrefine} & partial  & $\times$ & $\times$ & $\times$ \\
Reflexion \citep{reflexion}    & partial  & $\times$ & $\times$ & $\times$ \\
\textbf{ASP (this work)}       & \textbf{$\checkmark$} (@C, @S) & \textbf{$\checkmark$} (@G, A:[K,L,P,H]) & \textbf{$\checkmark$} (JSD, $\tau$) & \textbf{$\checkmark$} (Eq.~\ref{eq:routing}) \\
\bottomrule
\end{tabular}
\end{table}

\section{The Argent Signaling Protocol}
\label{sec:asp}

The core idea behind ASP is simple: before accepting any AI-generated answer, attach a machine-readable ``report card'' that rates the answer on multiple dimensions. A supervisor (human or automated) can read the card at a glance to decide whether to trust the answer, ask for a revision, or discard it. As in clinical lab reports, the header carries quality flags alongside the result.

\paragraph{Protocol scope vs.\ benchmark scope}
ASP is defined at the protocol level; this paper evaluates a subset of it empirically. The benchmarks in Sections~\ref{sec:expsetup}--\ref{sec:pipeline-results} exercise the header format (Section~\ref{sec:header}), the signal fields (Section~\ref{sec:signals}), the drift monitor (Section~\ref{sec:drift}), and the routing logic (Section~\ref{sec:routing}). Three protocol mechanisms are specified for completeness but are not the source of any routing decision reported in this paper: the stochasticity-modulated centroid update (Section~\ref{sec:drift}), the assumption decay and weight dampening (Section~\ref{sec:lifecycle}), and the K/L/P/H assumption classification (Section~\ref{sec:assumptions}). All three are designed for multi-turn deployments where assumptions accumulate over many exchanges; their empirical validation requires longer-horizon benchmarks and is future work (Section~\ref{sec:future-work}).

\subsection{System State}
At each orchestration step $t$, the controller tracks four pieces of structured state: a rolling conversation centroid $\vec{C}_t$ (a running average of where recent answers have been landing in quality space), the active assumption set $\mathbf{A}_t$, an adaptive decay rate $\lambda_t$ for assumption persistence, and per-agent error counters $\mathbf{E}_t$. ASP does not replace the natural-language payload; it exposes the minimal structured state needed to update these quantities deterministically, so every routing decision the controller makes can be reproduced from the telemetry log.

\subsection{Header Format}
\label{sec:header}
Every ASP-instrumented response carries a compact header alongside its natural-language content:

\smallskip
\noindent\texttt{[@C:X; @G:Y; @S:Z; A:[K:id, L:id, P:id, H:id]]}
\smallskip

\noindent Each signal field (@C, @G, @S) is a hexadecimal digit from 0 (lowest) to F (highest, i.e.\ 15), representing a quality score on a 16-point scale. The \texttt{A:[\ldots]} field is the assumption index, described in Section~\ref{sec:assumptions}. For example, a well-grounded answer that cites the right source but hedges slightly might carry:

\smallskip
\noindent\texttt{[@C:D; @G:F; @S:2; A:[K:42, L:09]]}
\smallskip

\noindent This tells the controller: certainty is high (D = 13/15), grounding is maximal (F = 15/15), stochasticity is low (2/15), and the answer relies on one known fact (K:42) and one learned derivation (L:09). Contrast this with an ungrounded answer:

\smallskip
\noindent\texttt{[@C:3; @G:1; @S:C; A:[H:17]]}
\smallskip

\noindent Here certainty is very low (3/15), grounding is nearly absent (1/15), stochasticity is high (C = 12/15), and the only assumption is hypothetical (H:17). The controller does not need to read the full answer text to know this response should not be passed downstream.

The header is designed to be:
\begin{itemize}
    \item \textbf{Compact}: a single line that can be parsed by any downstream consumer.
    \item \textbf{Auditable}: every field is mechanically derivable from the response and its context, with no opaque learned components.
    \item \textbf{Protocol-level}: the header travels with the response through any number of downstream agents, preserving provenance.
\end{itemize}

\subsection{Signal Fields}
\label{sec:signals}
The three signal fields capture complementary dimensions of response quality:
\begin{itemize}
    \item \textbf{@C (Certainty)}: how confident the response is in its own answer. High certainty means the response draws heavily on retrieved evidence and cites it correctly. Low certainty means the response relies on unsupported material or hedges extensively. In our benchmark, a Qwen answer that correctly cites the indemnification clause scores @C = F (maximum), while a SmolLM3 answer that produces uncited reasoning-mode paraphrase scores @C near 0.
    \item \textbf{@G (Grounding)}: how well the answer is anchored in retrieved evidence. This is the primary signal for distinguishing answers that are based on source material from those that are fabricated. A high @G score means most of the answer's content can be traced back to the evidence it was given. A @G of F means nearly every word in the answer appears in the source chunks; a @G of 0 means the answer shares almost no vocabulary with the source.
    \item \textbf{@S (Stochasticity)}: how much of the answer is unsupported or unstable. This is roughly the inverse of grounding, but weighted more heavily toward unsupported content and missing citations. A high @S score is a warning sign: the model is generating material that cannot be verified against the source. In our benchmark, SmolLM3's reasoning-mode paraphrases score @S in the 9-B range because they discuss the right topic but never cite the actual chunks.
\end{itemize}

\subsection{Estimating ASP Fields from Closed-API Agents}
\label{sec:closed-api}
ASP specifies the signal contract, not the estimator. The fields can be computed from white-box signals when available (logits, attention, retrieval traces) or from black-box outputs alone, making the protocol model-agnostic. The token-overlap estimator used in Section~\ref{sec:controller-results} is one such implementation chosen for transparency; here we outline three alternative estimators that require only text-level outputs and therefore work with hosted APIs (OpenAI, Anthropic, Google) that do not expose logits.

\paragraph{Certainty (@C) via verbalized confidence}
Following \citet{xiong_unc} and \citet{tian2023just}, we elicit verbalized confidence by appending a short instruction to the agent prompt: ``After your answer, output a single number in $[0,1]$ indicating your confidence that the answer is correct.'' The returned scalar is quantized to 4 bits for the ASP header. Calibration of verbalized confidence is imperfect but can be improved with prompt-level techniques \citep{tian2023just}; ASP treats the elicited value as an auditable proxy rather than a true posterior.

\paragraph{Grounding (@G) via NLI evidence coverage}
For each atomic claim $c_i$ extracted from the response and each retrieved chunk $r_j$, an NLI model produces an entailment score $s_{ij} = \mathrm{NLI}(r_j, c_i)$. Grounding is then $@G = |\{i : \max_j s_{ij} > \theta_g\}| / |\{c_i\}|$ with $\theta_g = 0.7$. Claim extraction can be done by a lightweight LLM call or by sentence segmentation; the NLI model itself need not match the generator.

\paragraph{Stochasticity (@S) via sample disagreement}
Drawing $k = 5$ completions at temperature $\tau = 0.7$ and computing semantic-cluster agreement following \citet{kuhn_unc} gives $@S = 1 - \text{agreement}$. This is the same semantic-entropy approach used in the calibration literature, adapted as a header field rather than a confidence target. None of these three estimators require logit access, retrieval-side hooks, or model fine-tuning: ASP can be deployed as a wrapper around any hosted API.

\paragraph{Non-validation disclaimer}
We do not validate these closed-API estimators empirically in this paper. The benchmark in Sections~\ref{sec:expsetup}--\ref{sec:pipeline-results} uses the token-overlap estimator described in Section~\ref{sec:signals} because it is fully transparent and reproducible for document-grounded QA. Empirical comparison of estimator families across white-box and black-box agents is a natural next step.

\subsection{Assumption Index}
\label{sec:assumptions}
Beyond the three numeric signals, ASP provides a structured \emph{assumption index} that classifies the evidentiary basis of each claim in the response. Every assumption referenced in the answer is tagged with one of four categories:

\begin{table}[t]
\centering
\caption{Assumption categories in the ASP header}
\label{tab:assumption-categories}
\begin{tabular}{clp{0.55\textwidth}}
\toprule
\textbf{Tag} & \textbf{Category} & \textbf{Meaning} \\
\midrule
K & Known & Established fact directly stated in the source document. The model is quoting or closely paraphrasing retrieved evidence. \\
L & Learned & Derived during processing (e.g., the model identified that a particular chunk contains the relevant definition). \\
P & Projected & Inferred from partial evidence. The source suggests this but does not state it directly. \\
H & Hypothetical & Speculative. No direct evidence supports this claim; the model is reasoning beyond the source. \\
\bottomrule
\end{tabular}
\end{table}

\paragraph{Trust gradient across categories}
The distinction tells the controller how much to trust each piece of the answer. A response built entirely on K (known) and L (learned) assumptions is citing and interpreting the actual document. A response that depends on P (projected) assumptions is extrapolating. A response built on H (hypothetical) assumptions is speculating. The controller can use this information to decide: repair is worth attempting when the answer is mostly K/L with some missing terms; containment is appropriate when the answer is mostly H with no grounding.

\paragraph{Assumption lifecycle (protocol-level design)}
The assumption index serves two purposes. First, it makes the evidentiary basis of each claim transparent to auditors: a response built on K~and~L assumptions is more trustworthy than one that depends on P~or~H. Second, in multi-turn deployments, it enables the controller to track each assumption through an \emph{active~$\to$~decayed~$\to$~quarantined} lifecycle: assumptions whose semantic vectors diverge from the running conversation centroid are exponentially down-weighted, and assumptions whose weights drop below a configurable floor are flagged as unreliable for any downstream consumer. The adaptive decay rate $\lambda_t$ and the type-dependent initial weight $W_0$ are specified in Section~\ref{sec:lifecycle}; the benchmark in this paper does not exercise the decay mechanism.

\paragraph{Operationalization status}
The assumption index is the least mature component of the protocol as evaluated here. In the current benchmark, routing decisions are driven entirely by the numeric signals (@C, @G, @S) and drift; the K/L/P/H labels are defined at the protocol level but are not extracted, labeled, or tracked per-response in the evaluation. A production implementation would need to specify how assumptions are identified (e.g., via claim extraction), how they are classified into K/L/P/H categories (e.g., by matching against retrieved evidence), and how labels are validated. These questions are open.

\subsection{Design Principles}
The signal fields use fixed-weight formulas rather than learned classifiers, so that any auditor can reproduce the scores from the raw response and context. The hexadecimal quantization sacrifices precision for compactness and readability. The assumption index uses four coarse categories rather than a continuous confidence spectrum, because the routing decisions it supports (repair vs.\ containment) are themselves coarse.

We prioritize interpretability over optimality: a controller that makes the right decision 85\% of the time and can explain each one is more useful in regulated settings (healthcare, finance, law) than one that makes the right decision 90\% of the time opaquely.

\subsection{Sidecar Deployment Model}
ASP is a separate interception layer that sits between agents rather than inside them. This is analogous to service mesh sidecars in microservices architectures (e.g., Envoy in Istio), which handle authentication, rate limiting, and observability without modifying the service itself.

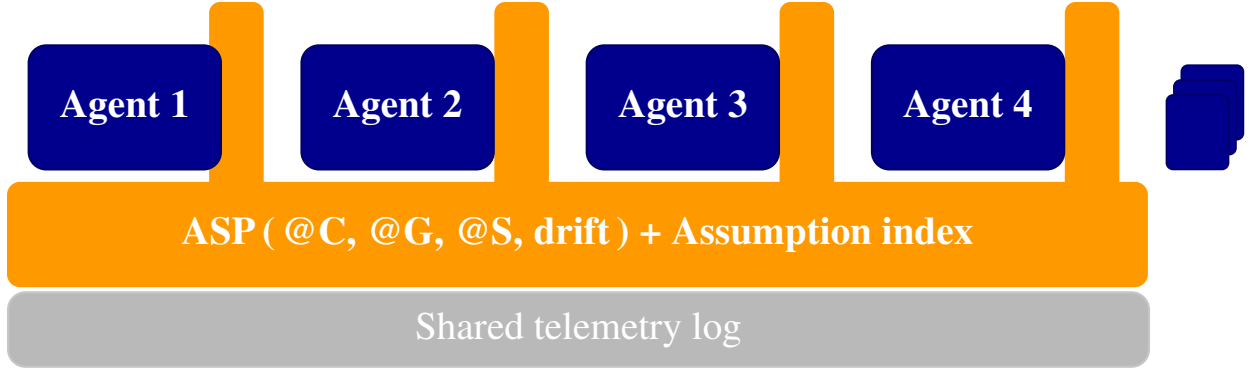
\begin{figure}[t]
\centering
\resizebox{\linewidth}{!}{%
\begin{tikzpicture}[
    x=1cm, y=1cm,
    agent/.style={fill=blue!55!black, draw=blue!40!black, line width=0.5pt, rounded corners=4pt,
                  minimum width=1.55cm, minimum height=1.0cm, align=center,
                  font=\footnotesize\bfseries, text=white},
    busfill/.style={fill=orange!80!yellow},
    telem/.style={fill=gray!55, draw=gray!45, line width=0.5pt, rounded corners=4pt,
                  minimum height=0.6cm, align=center, font=\footnotesize, text=white},
    logbox/.style={fill=blue!55!black, draw=blue!40!black, line width=0.4pt, rounded corners=1.5pt},
]
\fill[busfill, rounded corners=3pt] (-0.1, 0.1) rectangle (9.1, 0.95);

\foreach \x in {1.75, 4.05, 6.35, 8.65} {
    \fill[busfill, rounded corners=2.5pt] (\x-0.22, 0.85) rectangle (\x+0.22, 2.4);
}

\node[font=\footnotesize\bfseries, text=white] at (4.5, 0.525)
    {ASP\,(\,@C,\;@G,\;@S,\;drift\,) + Assumption index};

\node[agent] at (0.85, 1.55) {Agent 1};
\node[agent] at (3.05, 1.55) {Agent 2};
\node[agent] at (5.35, 1.55) {Agent 3};
\node[agent] at (7.65, 1.55) {Agent 4};

\foreach \i in {2,1,0} {
    \draw[logbox] (9.25+0.06*\i, 1.05+0.12*\i) rectangle ++(0.5, 0.6);
}

\node[telem, minimum width=9.2cm, anchor=south west] at (-0.1, -0.55) {Shared telemetry log};
\end{tikzpicture}
}
\caption{Sidecar-mediated architecture. The ASP layer (orange) is a single shared bus that all agents tap into via dedicated sidecars; the bus carries the @C/@G/@S signals, the drift measure, and the combined assumption index. Agents have no direct path to one another, so every inter-agent message is forced through the bus, which acts as the sole enforcement surface. All sidecar decisions are written to the shared telemetry log beneath the bus}
\label{fig:combs}
\end{figure}

Figure~\ref{fig:combs} illustrates the core architectural idea. The ASP layer is a single shared bus running beneath the agents, and each agent connects to it through a dedicated sidecar tap. An agent's only outbound channel is its tap into the bus: there is no direct agent-to-agent path. The bus computes @C/@G/@S, drift, and the combined assumption index, and is the sole enforcement surface for inter-agent traffic. Agents can be swapped, upgraded, or replaced without changing the bus logic, and vice versa.

\begin{figure}[t]
\centering
\includegraphics[width=\linewidth,trim=90 20 90 20,clip]{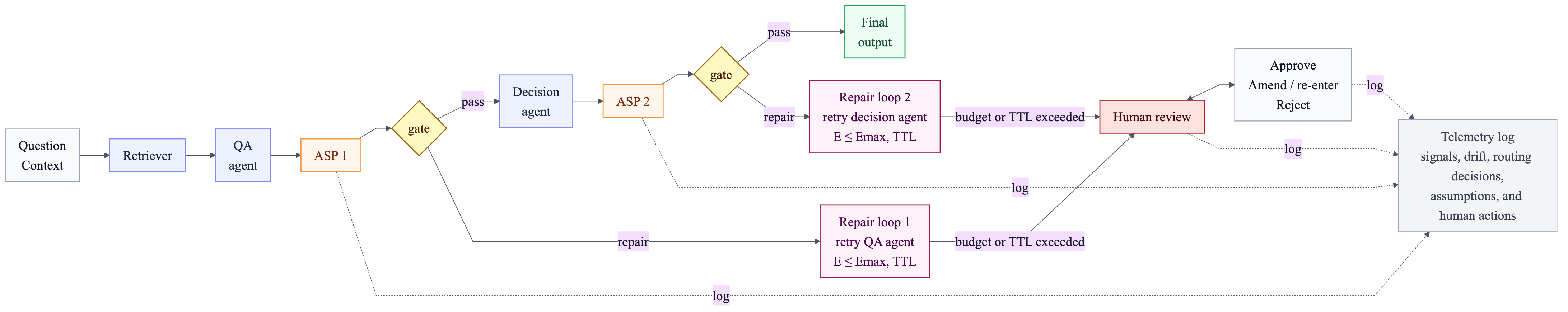}
\caption{Operational ASP pipeline. ASP sidecars gate each agent boundary, failed items enter bounded repair loops, unresolved cases escalate to human review, and telemetry records every decision}
\label{fig:pipeline-arch}
\end{figure}

Figure~\ref{fig:pipeline-arch} shows the operational pipeline that results from this interlocking design, including exception management, repair loops, and human escalation.

The architecture has four layers:

\paragraph{Main pipeline}
A question enters the retrieval agent, which provides context to Agent~A. Agent~A's output is intercepted by ASP Sidecar~1, which computes @C, @G, @S and drift and makes a deterministic gate decision. If passed, the output reaches Agent~B, whose output is similarly gated by Sidecar~2 before reaching the final output. Additional agents can be added to the chain; each boundary gets its own sidecar.

\paragraph{Repair loops}
When a sidecar triggers regeneration, the agent retries within its error budget ($E_{\max} = 2$ in our benchmark). Each retry is also bounded by a flow-level time-to-live (TTL) that prevents a pipeline from spinning indefinitely. The TTL is shared across all agents in a flow: if Agent~A uses two retries and Agent~B uses one, the flow has consumed three of its total budget.

\paragraph{Human review queue}
When the repair budget or TTL is exhausted, the item is routed to a human review queue rather than being silently dropped. A reviewer can approve the item as-is, amend the answer and re-enter it at the appropriate pipeline stage, or reject it entirely. Every human action is recorded in the telemetry log.

\paragraph{Telemetry}
Every sidecar decision (including the signal values that triggered it), every repair attempt, and every human action is logged. A compliance officer can reconstruct the complete decision history for any output without accessing the models or their internal state.

\medskip

This architecture means ASP operates in two modes within the same deployment:
\begin{itemize}
    \item \textbf{Standalone}: a single agent with an ASP controller loop (Section~\ref{sec:controller-results}).
    \item \textbf{Multi-agent}: ASP sidecars at every agent boundary, with repair loops, TTL enforcement, and human escalation (Section~\ref{sec:pipeline-results}).
\end{itemize}

\section{Controller Implementation}

This section describes how ASP signals are computed and used in the benchmark controller. The protocol itself (Section~3) defines \emph{what} signals are carried; this section defines \emph{how} they are derived and \emph{how} the controller acts on them. The same protocol could be implemented with different signal estimators (e.g., logit-based, verifier-based, or sampling-based); the benchmark uses a token-overlap estimator because it is transparent, reproducible, and sufficient for the document-grounded QA setting.

\subsection{Signal Computation}
The three signal values are not subjective judgments. They are computed mechanically by comparing the words in the AI's answer against the source material it was given. Figure~\ref{fig:signal-pipeline} shows the computation pipeline. For each generated draft, three raw features are extracted:

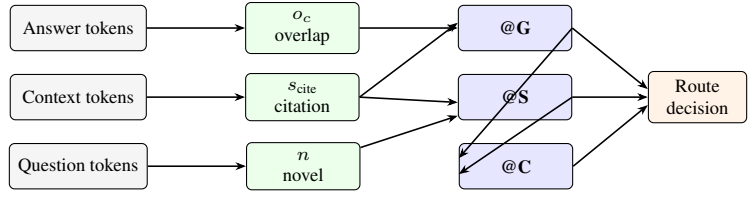
\begin{figure}[t]
\centering
\begin{tikzpicture}[
    node distance=0.4cm and 0.5cm,
    feat/.style={draw, rounded corners=2pt, fill=green!8, minimum width=1.5cm, minimum height=0.6cm, align=center, font=\scriptsize},
    sig/.style={draw, rounded corners=2pt, fill=blue!10, minimum width=1.5cm, minimum height=0.6cm, align=center, font=\scriptsize\bfseries},
    outcome/.style={draw, rounded corners=2pt, fill=orange!10, minimum width=1.3cm, minimum height=0.6cm, align=center, font=\scriptsize},
    inp/.style={draw, rounded corners=2pt, fill=gray!8, minimum width=1.8cm, minimum height=0.6cm, align=center, font=\scriptsize},
    arr/.style={-{Stealth[length=4pt]}, semithick},
]
\node[inp] (answer) {Answer tokens};
\node[inp, below=0.3cm of answer] (context) {Context tokens};
\node[inp, below=0.3cm of context] (question) {Question tokens};

\node[feat, right=1.3cm of answer] (oc) {$o_c$\\overlap};
\node[feat, right=1.3cm of context] (scite) {$s_{\text{cite}}$\\citation};
\node[feat, right=1.3cm of question] (n) {$n$\\novel};

\node[sig, right=1.3cm of oc] (G) {@G};
\node[sig, right=1.3cm of scite] (S) {@S};
\node[sig, right=1.3cm of n] (C) {@C};

\node[outcome, right=1.0cm of S] (route) {Route\\decision};

\draw[arr] (answer) -- (oc);
\draw[arr] (context) -- (scite);
\draw[arr] (question) -- (n);

\draw[arr] (oc) -- (G);
\draw[arr] (scite.east) -- ([yshift=2pt]G.west);
\draw[arr] (n) -- (S);
\draw[arr] (scite.east) -- ([yshift=-2pt]S.west);
\draw[arr] (G.east) -- ([yshift=3pt]C.west);
\draw[arr] (S.east) -- ([yshift=-3pt]C.west);

\draw[arr] (G.east) -- ([yshift=3pt]route.west);
\draw[arr] (S.east) -- (route.west);
\draw[arr] (C.east) -- ([yshift=-3pt]route.west);
\end{tikzpicture}
\caption{Signal computation pipeline. Raw features are extracted by token comparison; each signal draws on independent primary inputs. @G depends on overlap and citation, @S depends on novel ratio and citation, @C is the nonlinear product $@G \cdot (1 - @S)$}
\label{fig:signal-pipeline}
\end{figure}
Three raw features feed the signal fields (Equations~\ref{eq:grounding}--\ref{eq:certainty}):
\begin{itemize}
    \item $o_c$: \emph{context overlap}, the fraction of answer tokens that appear in any retrieved chunk. For example, if the model's answer contains 26 tokens and 24 of them appear somewhere in the three retrieved chunks, then $o_c = 24/26 = 0.92$.
    \item $s_{\text{cite}}$: \emph{citation score}, set to $1.0$ if the answer explicitly cites a valid retrieved chunk identifier (e.g., ``[Chunk~9]''), $0.4$ if it cites an identifier that was not in the retrieved set, and $0.0$ if it contains no citations at all.
    \item $n$ (\emph{novel\_ratio}): the fraction of answer tokens found in neither the retrieved context nor the question. Unlike unsupported ratio ($1 - o_c$), novel\_ratio excludes question-echo tokens (echoing the question is not fabrication).
\end{itemize}
A fourth feature is computed for use in the drift vector (Equation~\ref{eq:semvec}) but does not enter the signal fields directly:
\begin{itemize}
    \item $o_q$: \emph{question overlap}, the fraction of answer tokens that also appear in the question text. The drift vector treats high question overlap as a marker of answer profile. A well-grounded answer should add source information beyond echoing the question (see Section~\ref{sec:drift}).
\end{itemize}

The three ASP signal fields are then computed as follows:
\begin{align}
    @G &= \text{clamp}\bigl(0.60 \cdot o_c + 0.40 \cdot s_{\text{cite}}\bigr) \label{eq:grounding} \\
    @S &= \text{clamp}\bigl(0.70 \cdot n + 0.30 \cdot (1 - s_{\text{cite}})\bigr) \label{eq:stochasticity} \\
    @C &= @G \cdot (1 - @S) \label{eq:certainty}
\end{align}
where $\text{clamp}(x) = \min(1, \max(0, x))$ keeps each value between 0 and 1.

Table~\ref{tab:signal-design} summarizes the per-signal weight rationale, raw inputs, and high/low behavior.

\begin{table}[t]
\centering
\caption{Per-signal weight rationale, raw inputs, and high/low behavior. The features $o_c$ (context overlap), $s_{\text{cite}}$ (citation score), and $n$ (novel ratio) are defined in the bullet list above.}
\label{tab:signal-design}
\small
\begin{tabular}{>{\raggedright\arraybackslash}p{2.0cm}>{\raggedright\arraybackslash}p{4.5cm}>{\raggedright\arraybackslash}p{5.5cm}}
\toprule
\textbf{Parameter} & \textbf{Weight rationale} & \textbf{Signal independence and high / low behavior} \\
\midrule
\textbf{@G} \newline (Grounding) &
$0.60 \cdot o_c + 0.40 \cdot s_{\text{cite}}$. Context overlap weighted highest because using the source document's own language is the strongest evidence of document awareness. &
Depends on $o_c$ and $s_{\text{cite}}$; measures how well the answer is anchored in retrieved evidence. \textbf{High} when the answer mostly uses source words and cites the right section. \textbf{Low} when neither holds. \\
\addlinespace
\textbf{@S} \newline (Stochasticity) &
$0.70 \cdot n + 0.30 \cdot (1 - s_{\text{cite}})$. Novel content dominates because fabricated material is the primary risk in grounded QA. &
Depends on $n$ and $s_{\text{cite}}$; measures how much of the answer is material absent from both context and question. Because $n$ is not deterministically tied to $o_c$, @S captures a different failure mode from @G. \textbf{High} when the answer cannot be traced back to retrieved evidence (``making things up''). \textbf{Low} when novel content is rare and citation is valid. \\
\addlinespace
\textbf{@C} \newline (Certainty) &
$@G \cdot (1 - @S)$. Nonlinear product: high only when grounding is strong \emph{and} novelty is low, so @C cannot be high when either signal indicates a problem. &
Combines @G and @S into one complementary measure. \textbf{High} when both citation and overlap are strong (overall trustworthiness). \textbf{Low} when @G is weak or @S is high. \\
\bottomrule
\end{tabular}
\end{table}

The weights are hand-tuned rather than learned, which is a limitation (Section~\ref{sec:future-work}) but also a strength: any auditor can verify the computation from the raw token lists. Each value is compressed into a single hexadecimal digit (0~=~lowest, F~=~highest) for compact transmission in the ASP header.

\subsection{Drift Monitoring}
\label{sec:drift}
Individual signal values tell the controller about a single draft, but they do not reveal a trend. Is the model getting closer to a good answer with each retry, or is it wandering further away? Drift monitoring answers this question by measuring how far each answer's quality profile sits from what a well-grounded response typically looks like. A small drift means the answer is structurally similar to a good one (even if incomplete); a large drift means it has wandered into territory where refinement is unlikely to help.

\paragraph{Information-geometric motivation}
Information geometry motivates measuring movement on a statistical manifold through the Fisher information metric \citep{amari}. In practice, exact geodesic monitoring is too expensive for online orchestration. We therefore use Jensen--Shannon divergence (JSD) \citep{lin_jsd} as a bounded, symmetric surrogate. With base-2 logarithms, $\mathrm{JSD}(P \| Q) \in [0,1]$, and $\sqrt{\mathrm{JSD}}$ is a proper metric \citep{endres_metric}. This makes JSD attractive for auditable thresholding: the orchestrator gets a normalized drift signal without claiming access to an exact semantic geodesic.

\paragraph{Semantic vector and drift computation}
Each draft produces a four-element semantic vector that captures its quality profile:
\begin{equation}
    \mathbf{v} = \text{normalize}\bigl(\max(0.01, o_c),\; \max(0.01, s_{\text{cite}}),\; \max(0.01, n),\; \max(0.01, o_q)\bigr) \label{eq:semvec}
\end{equation}
Drift is the square root of the JSD between $\mathbf{v}$ and a fixed ideal QA reference vector $\mathbf{v}^* \propto (0.45, 0.35, 0.03, 0.17)$. This reference represents the quality profile of a well-grounded, well-cited answer: high context overlap (0.45), strong citation (0.35), low novel content (0.03), and moderate question overlap (0.17). The novel component is lower than the previous unsupported component because novel\_ratio is a stricter measure: tokens echoing the question are excluded, so a well-grounded answer should have very few truly novel tokens. Formally:
\begin{equation}
    d = \sqrt{\mathrm{JSD}(\mathbf{v} \,\|\, \mathbf{v}^*)} = \sqrt{\tfrac{1}{2}\bigl[\mathrm{KL}(\mathbf{v} \| \mathbf{m}) + \mathrm{KL}(\mathbf{v}^* \| \mathbf{m})\bigr]}, \quad \mathbf{m} = \tfrac{\mathbf{v} + \mathbf{v}^*}{2} \label{eq:drift}
\end{equation}

\paragraph{Selection of the ideal vector}
The ideal vector was hand-picked, not fit to held-out pass cases. The rationale is qualitative: a well-grounded QA answer should have high context overlap (the answer uses source language), strong citation (it references the right chunk), low novel content (little fabrication), and moderate question overlap (it adds information beyond echoing the question). The specific proportions $(0.45, 0.35, 0.03, 0.17)$ reflect that priority ordering. Because $\mathbf{v}^*$ together with the thresholds $\tau_{\text{warn}}$ and $\tau_{\text{crit}}$ influence routing decisions (particularly on SmolLM3, where containment vs.\ repair is determined per question), the paper owes a sensitivity analysis. We have not yet performed one. Section~\ref{sec:future-work} lists the needed ablations: perturbing $\mathbf{v}^*$ and $\tau_{\text{crit}}$ by $\pm 20\%$ to test whether the observed regime assignments survive, and computing per-model $\mathbf{v}^*$ rather than using a global reference.

\paragraph{Drift values in practice}
In our benchmark, a Qwen answer that cites the right chunk and uses the source document's language produces drift $d \approx 0.11$ (close to the ideal). A Dobby answer that cites correctly but misses some required terms produces drift $d \approx 0.20$ (moderate). A SmolLM3 answer that paraphrases the topic without citations produces $d \approx 0.62$ (far from the ideal). These values are interpretable because JSD is bounded and symmetric: 0 means identical to the ideal, 1 means maximally different.

\paragraph{Centroid tracking (designed for longer trajectories)}
The controller also maintains a rolling conversation centroid $\mathbf{c}$ that tracks where the agent's recent answers have been landing in semantic-vector space. The update applies a blend coefficient that shrinks for highly stochastic drafts, so that noisy outputs contribute less to the running estimate:
\begin{equation}
    \beta_{\text{eff}} = \text{clamp}\bigl(\beta \cdot \max(0.05,\; 1 - @S/30)\bigr), \quad \mathbf{c} \leftarrow \text{normalize}\bigl((1 - \beta_{\text{eff}}) \cdot \mathbf{c} + \beta_{\text{eff}} \cdot \mathbf{v}\bigr) \label{eq:centroid}
\end{equation}
with $\beta = 0.35$. In the current benchmark, the centroid operates over only 1--3 drafts per question, so the routing decisions reported in Section~\ref{sec:controller-results} are driven primarily by per-draft drift $d$ against the fixed ideal vector $\mathbf{v}^*$, not by centroid-based trajectory tracking. Centroid-based routing is expected to matter more in multi-turn settings where an agent's quality may shift gradually over dozens of exchanges.

\paragraph{Sensitivity to miscalibrated confidence signals}
A reasonable concern is what happens when the elicited certainty signal is itself miscalibrated. Let the observed certainty satisfy $@C_{\text{obs}} = @C_{\text{true}} + \varepsilon$, with $\varepsilon$ having mean $\mu$ and standard deviation $\sigma$. Two regimes matter for the orchestrator. \emph{Bias regime} ($\mu \neq 0$): because routing decisions in Equation~\ref{eq:routing} depend on $\Delta\mathrm{JSD}$ relative to the running centroid (Equation~\ref{eq:centroid}), a bias shared across agents cancels in the JSD computation and does not affect control flow; per-agent bias is detectable through cross-agent disagreement and can be debiased online. \emph{Variance regime} ($\sigma > 0$): near-threshold flips occur with probability $\Phi\!\bigl((\tau - |\mathrm{JSD} - \mu|)/\sigma\bigr)$, where $\Phi$ is the standard normal CDF. Using $\sigma \approx 0.05$ (the empirical magnitude reported by \citet{tian2023just} for verbalized confidence on Llama-class models) gives a near-threshold flip rate of roughly 5\%. We mitigate this with hysteresis: a payload must exceed $\tau_{\text{crit}}$ for two consecutive turns before quarantine is triggered. This makes the controller robust to shared bias outright and to per-agent variance within plausible empirical magnitudes.

\subsection{Assumption Lifecycle Tracking}
\label{sec:lifecycle}
The controller also tracks each assumption referenced in a response through an \emph{active~$\to$~decayed~$\to$~quarantined} lifecycle. When the controller detects that an assumption's semantic vector is diverging from the rolling conversation centroid, the assumption's weight is exponentially dampened. The decay rate $\lambda_t$ is itself adaptive, computed from the rolling variance of recent drift measurements:
\begin{equation}
\lambda_t = \alpha \cdot \mathrm{Var}(\mathrm{JSD}_{t-k \dots t}) + (1-\alpha)\lambda_{t-1}, \label{eq:decay-rate}
\end{equation}
and the effective weight of assumption $A_i$ at time $t$ is
\begin{equation}
W(A_i, t) = W_0 \cdot \exp\bigl(-\lambda_t \cdot \mathrm{JSD}(\vec{A}_i \,\|\, \vec{C}_t)\bigr), \label{eq:assumption-weight}
\end{equation}
where the initial weight $W_0$ is type-dependent: known (K) and learned (L) assumptions start with higher weight than projected (P) or hypothetical (H) ones. When $W(A_i, t)$ drops below a configurable floor (0.18 in the protocol specification), the assumption is marked \textbf{decayed}. If the controller yields due to unrecoverable drift, all active assumptions are \textbf{quarantined}, flagging them as unreliable for any downstream consumer. As with centroid tracking, the benchmark trajectories in this paper (1--3 drafts per question) are too short to exercise the decay mechanism; it is included here as part of the controller specification for multi-turn deployments.

\subsection{Routing Logic}
\label{sec:routing}
Given the signal values and drift score, the controller applies a deterministic routing policy:
\begin{equation}
R(\text{agent}) =
\begin{cases}
\text{Execute}, & \text{if } E < E_{\max} \land d \le \tau_{\text{warn}} \land g \geq g_{\min} \land s \leq \sigma_{\max} \\
\text{Warn}, & \text{if } E < E_{\max} \land \tau_{\text{warn}} < d \le \tau_{\text{crit}} \\
\text{Regenerate}, & \text{if } E < E_{\max} \land (d > \tau_{\text{crit}} \lor g < g_{\min} \lor s > \sigma_{\max}) \\
\text{Yield}, & \text{if } E \ge E_{\max}.
\end{cases}
\label{eq:routing}
\end{equation}
Here $g$ and $s$ are the raw (continuous) grounding and stochasticity values from Equations~\ref{eq:grounding}--\ref{eq:stochasticity}, not the quantized hex digits. The controller operates on these raw floats; hex quantization is a serialization convention for the ASP header, not the internal decision surface. ASP signals drive all routing decisions in a unified policy: grounding, stochasticity, and drift jointly determine whether a draft is executed, repaired, or contained. In this benchmark, $\tau_{\text{warn}} = 0.12$, $\tau_{\text{crit}} = 0.22$, $g_{\min} = 0.33$, $\sigma_{\max} = 0.67$, and $E_{\max} = 2$, giving each question at most three drafts: the initial answer plus up to two controller-triggered retries. Unlike common retry loops, each regeneration is explicitly logged: a fresh ASP header is emitted, the assumption index is updated, and the telemetry records which traces were rejected and why.

\subsection{Failure Classification}
The key design choice in this benchmark is to separate two kinds of failure that are usually lumped together. When a student turns in a partially correct exam answer, a teacher can see whether the student understood the material but ran out of time (repairable) or guessed randomly (not worth correcting). The controller makes the same distinction for AI answers.

An answer \emph{passes} if it cites supporting retrieved evidence and covers at least 60\% of the required terms. Failing answers are classified into failure regimes based on their grounding profile:
\begin{itemize}
    \item \textbf{Grounded partial}: the answer matches a valid citation and overlaps the correct material, but does not cover enough required terms. The controller targets these for \emph{repair}. In our benchmark, these are answers where the model cited the right section but missed some branches of a multi-part definition.
    \item \textbf{Citation gap}: the answer overlaps the right material (context overlap $\geq 0.65$ and term coverage $\geq 0.25$) but cites the wrong chunk or no chunk. Also targeted for \emph{repair}. These are answers that use the right vocabulary but fail to attribute it to the correct source.
    \item \textbf{Ungrounded}: the answer lacks sufficient grounding to justify refinement. The controller targets these for \emph{containment}. In our benchmark, these are primarily SmolLM3's reasoning-mode paraphrases that discuss the right topic but never cite a source.
\end{itemize}

Table~\ref{tab:failure-classes} summarizes the failure classification.

\begin{table}[t]
\centering
\caption{Failure classes used by the controller benchmark}
\label{tab:failure-classes}
\begin{tabular}{lll}
\toprule
\textbf{Condition} & \textbf{Class} & \textbf{Controller target} \\
\midrule
Citation match and term coverage $\geq 0.60$ & pass & execute or warn \\
Citation match, coverage $< 0.60$ & grounded partial & repair \\
Overlap $\geq 0.65$, coverage $\geq 0.25$, wrong cite & citation gap & repair \\
Insufficient grounding for refinement & ungrounded & containment \\
\bottomrule
\end{tabular}
\end{table}

\subsection{Parameter Selection}
\label{sec:params}
ASP has multiple hand-tuned parameters: the linear weights in Equations~\ref{eq:grounding}--\ref{eq:stochasticity}, the ideal vector $\mathbf{v}^* \propto (0.45, 0.35, 0.03, 0.17)$ in Section~\ref{sec:drift}, the routing thresholds $\tau_{\text{warn}} = 0.12$, $\tau_{\text{crit}} = 0.22$, $g_{\min} = 0.33$, $\sigma_{\max} = 0.67$, the centroid blend $\beta = 0.35$, the assumption decay floor $0.18$, the citation score levels $\{0.0, 0.4, 1.0\}$, the retrieval definition bonuses $\{+0.5, +1.0\}$, and the error budget $E_{\max} = 2$. All values were set before benchmark execution based on the qualitative rationale stated in the corresponding sections: grounding weighted highest in $@G$ (the answer should use source language), novel content weighted highest in $@S$ (fabrication is the primary risk), and the well-grounded answer profile encoded in $\mathbf{v}^*$. No parameter was adjusted in response to benchmark results. Sensitivity to these choices is the primary target for follow-on ablation work (Section~\ref{sec:future-work}).

\section{Experimental Setup}
\label{sec:expsetup}

\subsection{Task and Document}
To test the controller in a realistic setting, we use a publicly available pharmaceutical license agreement between Array BioPharma and Ono Pharmaceutical (a real-world legal document with precise definitions, cross-references, and multi-clause provisions that challenge AI systems). Legal agreements are a good stress test for grounded QA because they contain exactly the kind of multi-branch definitions, conditional provisions, and cross-referenced terms that expose both retrieval failures and generation failures. The task is a 27-question document-grounded QA benchmark over this agreement. Questions span seven categories covering the major functional areas of the agreement, as shown in Table~\ref{tab:question-categories}.

\begin{table}[t]
\centering
\caption{Question categories in the 27-question Array/Ono benchmark}
\label{tab:question-categories}
\begin{tabular}{lr}
\toprule
\textbf{Category} & \textbf{Count} \\
\midrule
Definitions (Change in Control, Field, First Commercial Sale, etc.) & 7 \\
License scope (development, manufacturing, sublicense, grant-back) & 6 \\
Termination and wind-down & 5 \\
General provisions (governing law, amendment, assignment, etc.) & 5 \\
Governance (JDRC establishment and duties) & 2 \\
Risk allocation (recall control, indemnification) & 2 \\
\bottomrule
\end{tabular}
\end{table}

Each question specifies an expected answer, a set of 3--4 required terms that must appear for full coverage, and citation references (section numbers with line spans) that identify the authoritative source text. For example, the question ``What counts as a Change in Control?'' requires the terms ``fifty percent,'' ``majority,'' ``merger,'' and ``substantially all,'' and expects a citation to Section~1.6 of the agreement.

\subsection{Retrieval}
Before the model can answer a question, the system must find the relevant sections of the agreement. Evidence is retrieved using TF-IDF (term frequency--inverse document frequency), a standard statistical method that scores relevance by how often query words appear in each passage relative to the rest of the document. The agreement text is stored in a portable vector store. The document is chunked at approximately 220 words per chunk with one-paragraph overlap, producing a set of passages that each preserve their line span, article heading, and section reference.

At query time, the store computes sparse dot-product similarity between the query vector and all chunk vectors, with definition-aware bonuses that help locate formal definitions: $+1.0$ for chunks containing an explicit ``\emph{X} means'' pattern (the standard legal definition syntax), $+0.5$ for other definition syntax, and smaller bonuses for chunks in the Article I definitions block. The top-$k=3$ chunks are returned as the grounding context for generation.

This sparse retrieval approach is sufficient for most single-section definitions but can miss multi-branch clauses that span non-adjacent paragraphs. For instance, the ``Change in Control'' definition includes branches about voting stock, board control, mergers, and asset sales, which may span multiple chunks. Domain-aware embedding retrofitting \citep{sharma_retrofit} could improve recall in such cases by learning which terms should be retrieved together even when they appear in different paragraphs.

\subsection{Generation and Evaluation}
Each question is answered with a maximum token budget of 96 tokens. The prompt instructs the model to answer concisely and cite supporting chunk identifiers. Temperature is set to $0.1$ on the first attempt and $0.05$ on retries to reduce variance while still allowing some generation diversity.

We compare two conditions:
\begin{itemize}
    \item \textbf{Baseline}: one grounded answer attempt with no ASP-driven intervention. The model sees the question and the top-3 retrieved chunks and produces a single answer.
    \item \textbf{ASP controller}: the same initial generation followed by controller assessment. The controller computes @C, @G, @S, and drift, then decides whether to accept, warn, regenerate, or yield. If regenerated, the model is prompted again with feedback about what was missing.
\end{itemize}

An answer \emph{passes} if two conditions are met: 
\begin{list}{answer-conditions}{spacing}
    \item The answer cites at least one chunk that matches the expected citation for the question.
    \item The fraction of required terms present in the answer (case-insensitive) is at least $0.60$.
\end{list}

This is a deliberately strict rubric: an answer that discusses the right topic but cites the wrong section or misses key terms still fails. 
The strictness is intentional because it forces the controller to demonstrate measurable improvement on well-defined criteria rather than on subjective quality judgments.

\subsection{Models}
The benchmark evaluates three language models running locally on consumer hardware. Each model is stored in the GGUF format, a compact binary format that allows large language models to run on laptops and desktops without cloud infrastructure. The three models were chosen to span a range of capability levels and to produce qualitatively different failure modes:
\begin{itemize}
    \item \texttt{Qwen3.5-0.8B-BF16}: a compact 0.8B-parameter instruction-tuned model. Despite its small size, Qwen follows instructions well and produces grounded answers that reference the source material. Its failures tend to be incomplete rather than fabricated.
    \item \texttt{dobby-8b-unhinged-q6\_k}: an 8B-parameter model not instruction-tuned for document QA. Despite lacking task-specific instruction tuning, Dobby produces coherent grounded answers on a meaningful fraction of questions, with failures tending toward incomplete coverage rather than fabrication. This makes it a useful test case for repair alongside Qwen.
    \item \texttt{SmolLM3-3B-Q8\_0}: a 3B-parameter model with moderate instruction following. SmolLM3 sometimes produces grounded answers and sometimes drifts into uncited reasoning-mode paraphrases, making it the ideal test case for per-question triage.
\end{itemize}

Qwen and Dobby were run using MLX, Apple's optimized framework for running AI models on Mac hardware. SmolLM3 required a different runtime (\texttt{llama.cpp}, a widely used open-source inference engine) because the MLX loader could not initialize that particular model file. This keeps within-model baseline-versus-ASP comparisons usable, but it weakens any strong cross-model interpretation that assumes backend-invariant behavior.

The primary metrics are pass rate, citation match rate, mean required-term coverage, yield rate, and controller action counts. The benchmark also reports controller confusion matrices, but those should be interpreted carefully: they measure agreement between controller actions and benchmark-induced repair/containment labels, not independent estimates of optimal policy accuracy. We therefore treat action counts and end-task metrics as the primary evidence.

\section{Results: Standalone Controller}
\label{sec:controller-results}

Table~\ref{tab:model-results} reports per-model outcomes on the 27-question benchmark. Two patterns are immediate: the controller does not uniformly improve end-task success, and the same controller serves different operational roles depending on the base model.

\begin{table}[t]
\centering
\caption{Per-model outcomes on the 27-question Array/Ono benchmark. Rates are percentages. Action profile reports repair interventions (R), containment interventions (C), and clean pass-throughs (P)}
\label{tab:model-results}
\begin{tabular}{lccccll}
\toprule
\textbf{Model} & \textbf{Backend} & \textbf{Pass} & \textbf{Cite} & \textbf{Coverage} & \textbf{Actions} & \textbf{Observed role} \\
\midrule
Qwen 0.8B & MLX & 11.1 $\rightarrow$ 33.3 & 55.6 $\rightarrow$ 59.3 & 36.7 $\rightarrow$ 65.4 & 24R / 0C / 3P & Repair layer \\
Dobby 8B & MLX & 33.3 $\rightarrow$ 44.4 & 70.4 $\rightarrow$ 66.7 & 48.8 $\rightarrow$ 54.3 & 19R / 0C / 8P & Repair layer \\
SmolLM3 3B & llama.cpp & 0.0 $\rightarrow$ 0.0 & 18.5 $\rightarrow$ 18.5 & 24.7 $\rightarrow$ 25.6 & 15R / 12C / 0P & Mixed triage \\
\bottomrule
\end{tabular}
\end{table}

\noindent Note that Dobby's baseline pass rate of 33.3\% is notable for a model not instruction-tuned for document QA. Despite being 8B parameters, Dobby produces coherent grounded answers on a meaningful fraction of questions. The ASP controller adds 4 fail-to-pass recoveries (with 1 regression), raising the pass rate to 44.4\%.

\paragraph{Aggregate view}
Across all three models, the benchmark covers 81 model-question pairs. Aggregate passes move from 12/81 in baseline to 21/81 under ASP, with 10 fail-to-pass recoveries and 1 regression. The controller applies 58 repair interventions, 12 containment interventions, and 11 clean pass-throughs. The aggregate numbers support a meaningful reading: ASP produces measurable gains while adapting its operational role to each model's failure regime. The protocol's value lies in both failure differentiation and task improvement.

\subsection{Qwen: Repair Layer}
Qwen is the strongest example of ASP as a repair controller. Under the benchmark evaluator, Qwen produces zero ungrounded failures. All non-passing answers are anchored in the correct agreement material; they fail because they omit required branches, qualifiers, or exact citations. The controller classifies all failures as repairable and attempts regeneration on each. Three questions pass on the initial draft and are routed through without intervention.

ASP produces 6 fail-to-pass recoveries with 0 regressions, raising the pass rate from 11.1\% to 33.3\%. Citation match rises from 55.6\% to 59.3\%, and mean term coverage rises from 36.7\% to 65.4\%. The 29-point improvement in term coverage is particularly notable: it means that on average, ASP-instrumented answers include substantially more of the required legal terms than baseline answers. The action profile (24R / 0C / 3P) confirms that the controller operates entirely in repair mode, with no containment interventions.

The best interpretation is that ASP helps Qwen as a structured reviewer/editor: it pushes drafts toward more complete, better-cited answers. The 0\% ungrounded rate means every failure is amenable to repair, and the controller exploits this by consistently attempting regeneration rather than yielding.

\subsection{Dobby: Repair Layer}
Dobby illustrates that model size and instruction tuning are not the only determinants of grounded QA performance. Despite not being instruction-tuned for document QA, the 8B-parameter Dobby produces coherent grounded answers on a meaningful fraction of questions, achieving a 33.3\% baseline pass rate with 70.4\% citation match and 48.8\% mean term coverage.

The ASP controller operates primarily in repair mode (19R / 0C / 8P), producing 4 fail-to-pass recoveries and 1 regression. Pass rate rises from 33.3\% to 44.4\%, and mean term coverage improves from 48.8\% to 54.3\%. Citation match decreases slightly from 70.4\% to 66.7\%, suggesting that some repair attempts broaden coverage at the cost of citation precision. Dobby maintains 0\% ungrounded failures, meaning every non-passing answer is anchored in the correct agreement material.

The failure profile for non-passing answers splits between grounded partial failures (37.0\%) and citation gaps (33.3\%). These are structurally similar to Qwen's failure modes: the model identifies the right material but omits required branches or cites imprecisely. The 8 clean pass-throughs (questions that pass on the initial draft without intervention) confirm that Dobby produces baseline-quality answers on nearly a third of questions. ASP adds value by converting 4 additional borderline failures into passes through targeted repair.

\subsection{SmolLM3: Mixed Triage}
SmolLM3 is the primary mixed-triage case, with both repair (15) and containment (12) interventions in the action profile (15R / 12C / 0P). End-task success does not improve (0.0\% baseline and ASP pass rate), and citation match holds steady at 18.5\%. Mean term coverage edges up slightly from 24.7\% to 25.6\%. The ungrounded rate is 44.4\%.

This regime is important because it shows that ASP need not be globally optimistic or globally pessimistic. On some questions, SmolLM3 stays attached to the correct definition chunk with reasonable grounding scores (@G in the C-E range) and the controller legitimately attempts repair. On other questions, the same model drifts into uncited reasoning-mode paraphrase (@G dropping, @S rising, drift increasing) and the controller correctly routes to containment. The same controller can dispatch between repair and halt at the per-question level based on the signal profile of each individual draft. Even when final benchmark success remains poor, the routing behavior is materially different from treating every failure as just another retry opportunity.

\section{Signal Trajectory Analysis}

This section presents four representative controller trajectories formatted as before/after comparisons. Each card shows the question, the baseline response, and the ASP controller outcome with per-attempt signal values. Required terms for each question are listed in the header; matched terms appear in \textbf{bold} in the response text. These examples make the abstract signal values concrete by showing exactly what the model said, what the controller measured, and what action was taken.

\medskip

\begin{trajectorycard}{Q: When does Ono indemnify Array? \hfill \normalfont\small Qwen 0.8B (repair $\rightarrow$ \textbf{pass})}
\small
\textit{Required terms:} Third Party Claim, Ono Territory, clinical studies, gross negligence or willful misconduct

\smallskip
\colorbox{qlight}{\parbox{\dimexpr\linewidth-2\fboxsep}{%
\textbf{Baseline Response} \hfill \texttt{@G=F \enspace @S=0 \enspace d=0.176 \enspace cov=50\%}\\[3pt]
Ono shall indemnify Array against \textbf{Third Party Claims} arising from the use, marketing, distribution, or sale of Array Products, injury to patients in \textbf{Clinical Studies}, or breaches of representations and warranties by Array, as defined in Article~15 [Chunk~281].
}}

\smallskip
\colorbox{qlight}{\parbox{\dimexpr\linewidth-2\fboxsep}{%
\textbf{After ASP Controller} \hfill \texttt{@G=F \enspace @S=1 \enspace d=0.278 \enspace cov=75\%}\\[3pt]
Ono shall indemnify Array against \textbf{Third Party Claims} arising from the use, marketing, distribution, importation, or sale of Array Products, injury to patients in \textbf{Clinical Studies}, or breaches of representations and warranties by Array, as defined in Article~15 [Chunk~281]. Additionally, Ono shall indemnify Array against claims resulting from the \textbf{gross negligence or willful misconduct} of an Array Indemnitee [Chunk~280].
}}

\smallskip
\textbf{Outcome:} PASS. Repair added Chunk~280 (exception clause), recovering one missing required term. ``Ono Territory'' was never recovered.
\end{trajectorycard}

\medskip

\begin{trajectorycard}{Q: What counts as a Change in Control? \hfill \normalfont\small Qwen 0.8B (repair $\rightarrow$ \textbf{yield})}
\small
\textit{Required terms:} fifty percent, majority, merger, substantially all

\smallskip
\colorbox{qlight}{\parbox{\dimexpr\linewidth-2\fboxsep}{%
\textbf{Baseline Response} \hfill \texttt{@G=D \enspace @S=0 \enspace d=0.107 \enspace cov=25\%}\\[3pt]
A Change in Control is defined as any event occurring after the Effective Date where a party becomes the beneficial owner of more than 50\% of outstanding Voting Stock or has the power to elect a \textbf{majority} of the board of directors [Chunk~9].
}}

\smallskip
\colorbox{qlight}{\parbox{\dimexpr\linewidth-2\fboxsep}{%
\textbf{After ASP Controller (attempt 2)} \hfill \texttt{@G=E \enspace @S=1 \enspace d=0.112 \enspace cov=50\%}\\[3pt]
A Change in Control under this agreement is defined as any of the following events occurring after the Effective Date: (1)~a party becomes the beneficial owner of more than \textbf{fifty percent} of outstanding Voting Stock or (2)~a party has the power to elect a \textbf{majority} of the board of directors [Chunk~9].
}}

\smallskip
\textbf{Outcome:} YIELD after 2 regenerations. Grounding stayed high ($@G \geq \text{D}$, drift $\leq 0.112$), confirming the answer was repairable. ``merger'' and ``substantially all'' were absent from the top-3 retrieved chunks, making this a retrieval limitation rather than a generation failure.
\end{trajectorycard}

\medskip

\begin{trajectorycard}{Q: What counts as a Change in Control? \hfill \normalfont\small SmolLM3 3B (containment $\rightarrow$ \textbf{yield})}
\small
\textit{Required terms:} fifty percent, majority, merger, substantially all

\smallskip
\colorbox{qlight}{\parbox{\dimexpr\linewidth-2\fboxsep}{%
\textbf{Baseline Response} \hfill \texttt{@G=3 \enspace @S=A \enspace d=0.620 \enspace cov=0\%}\\[3pt]
$\langle$think$\rangle$ \ldots\ A Change in Control would typically involve a shift in ownership or management\ldots\ [reasoning-mode paraphrase, no chunk citations, no source terms]
}}

\smallskip
\colorbox{qlight}{\parbox{\dimexpr\linewidth-2\fboxsep}{%
\textbf{After ASP Controller (attempt 3)} \hfill \texttt{@G=2 \enspace @S=B \enspace d=0.650 \enspace cov=0\%}\\[3pt]
$\langle$think$\rangle$ \ldots\ Change in Control generally refers to an acquisition or transfer of control\ldots\ [continued reasoning-mode paraphrase, no chunk citations]
}}

\smallskip
\textbf{Outcome:} YIELD. Signals degraded across attempts: $@G$ dropped 3$\rightarrow$2, $@S$ rose A$\rightarrow$B, drift increased 0.620$\rightarrow$0.650. The model stayed in uncited reasoning mode, producing plausible-sounding text with no connection to the retrieved chunks. Correctly routed to containment.
\end{trajectorycard}

\medskip

\begin{trajectorycard}{Q: What is a First Commercial Sale? \hfill \normalfont\small SmolLM3 3B (containment $\rightarrow$ \textbf{yield})}
\small
\textit{Required terms:} bona fide, arm's length, Ono Territory, Marketing Approval

\smallskip
\colorbox{qlight}{\parbox{\dimexpr\linewidth-2\fboxsep}{%
\textbf{Baseline Response} \hfill \texttt{@G=7 \enspace @S=9 \enspace d=0.578 \enspace cov=50\%}\\[3pt]
$\langle$think$\rangle$ \ldots\ The first line there says, ``1.20 `First Commercial Sale' means\ldots'' [reasoning-mode paraphrase, no chunk citations]
}}

\smallskip
\colorbox{qlight}{\parbox{\dimexpr\linewidth-2\fboxsep}{%
\textbf{After ASP Controller (attempt 3)} \hfill \texttt{@G=6 \enspace @S=B \enspace d=0.610 \enspace cov=0\%}\\[3pt]
$\langle$think$\rangle$ \ldots\ The definition of First Commercial Sale is\ldots\ [continued reasoning-mode paraphrase, no chunk citations]
}}

\smallskip
\textbf{Outcome:} YIELD. Signals degraded monotonically: $@G$ dropped 7$\rightarrow$6, $@S$ rose 9$\rightarrow$B, drift increased 0.578$\rightarrow$0.610. The model stayed in uncited paraphrase mode and was correctly routed to containment.
\end{trajectorycard}

\medskip

The four trajectories illustrate the controller's adaptive behavior. On the same benchmark, Qwen's high grounding ($@G \geq \text{D}$) and low stochasticity ($@S \leq 1$) kept it in the repair regime. SmolLM3's intermediate signals ($@G$ in 3--7, $@S$ in 9--B) show the transition zone where the controller must decide per-question whether to repair or halt.

\section{Results: Multi-Agent Pipeline}
\label{sec:pipeline-results}

The standalone controller benchmark (Section~\ref{sec:controller-results}) demonstrates ASP within a single agent's retry loop. To test whether ASP prevents semantic drift \emph{across agent boundaries}, we evaluate a two-agent pipeline where an upstream retrieval agent (Agent~A) answers a question and a downstream decision agent (Agent~B) must make a risk assessment based on that answer.

\subsection{Pipeline Design}
Agent~A is the same retrieval QA agent from the controller benchmark: it receives a question and top-$k=3$ chunks and produces a grounded answer. Agent~B receives Agent~A's output and must classify the provision as HIGH RISK, MODERATE RISK, or LOW RISK with a one-sentence justification grounded in the specific terms cited. We compare three conditions:

\begin{itemize}
    \item \textbf{Baseline}: Agent~B receives Agent~A's raw text with no quality metadata. If Agent~A hallucinated, Agent~B has no structured way to detect it.
    \item \textbf{ASP gated}: The ASP sidecar intercepts Agent~A's output. If grounding falls below $g_{\min}$, stochasticity exceeds $\sigma_{\max}$, or drift exceeds $\tau_{\text{crit}}$, the output is \emph{blocked} and Agent~B receives an escalation message instead.
    \item \textbf{ASP visible}: Agent~B receives Agent~A's output together with the full ASP header, so it can reason about upstream quality in its own judgment.
\end{itemize}

\begin{figure}[t]
\centering
\begin{tikzpicture}[
    node distance=0.4cm and 0.6cm,
    agent/.style={draw, rounded corners=3pt, fill=blue!8, minimum width=1.4cm, minimum height=0.7cm, align=center, font=\scriptsize},
    sidecar/.style={draw, rounded corners=3pt, fill=orange!12, minimum width=1.4cm, minimum height=0.7cm, align=center, font=\scriptsize},
    block/.style={draw, rounded corners=3pt, fill=red!10, minimum width=1.4cm, minimum height=0.7cm, align=center, font=\scriptsize},
    passn/.style={draw, rounded corners=3pt, fill=green!10, minimum width=1.4cm, minimum height=0.7cm, align=center, font=\scriptsize},
    lbl/.style={font=\small\bfseries, anchor=south},
    arr/.style={-{Stealth[length=4pt]}, semithick},
    xarr/.style={-{Stealth[length=4pt]}, semithick, red!60!black},
]
\node[lbl] at (1.8, 2.2) {Baseline (no ASP)};
\node[agent] (a1) at (0, 1.4) {Agent A};
\node[agent] (b1) at (3.6, 1.4) {Agent B};
\node[font=\scriptsize, align=center] (mid1) at (1.8, 1.4) {27 answers\\(14 ungrounded)};
\draw[arr] (a1) -- (mid1);
\draw[arr] (mid1) -- (b1);
\node[font=\scriptsize, color=red!70!black, align=center] at (1.8, 0.6) {51.8\% ungrounded\\content propagated};

\node[lbl] at (8.0, 2.2) {ASP Gated};
\node[agent] (a2) at (5.6, 1.4) {Agent A};
\node[sidecar] (sc) at (8.0, 1.4) {ASP\\Sidecar};
\node[agent] (b2) at (10.4, 1.4) {Agent B};
\draw[arr] (a2) -- node[above, font=\tiny] {27} (sc);
\draw[arr] (sc) -- node[above, font=\tiny] {3} (b2);
\node[block] (blk) at (8.0, 0.3) {Blocked: 24};
\draw[xarr] (sc) -- (blk);
\node[font=\scriptsize, color=green!50!black, align=center] at (8.0, -0.4) {0\% ungrounded\\content propagated};
\end{tikzpicture}
\caption{Pipeline comparison. \textbf{Left}: without ASP, all 27 outputs (including 14 ungrounded) reach Agent~B. \textbf{Right}: the ASP sidecar blocks 24 outputs, allowing only 3 grounded answers through. Zero ungrounded content reaches Agent~B}
\label{fig:pipeline-compare}
\end{figure}
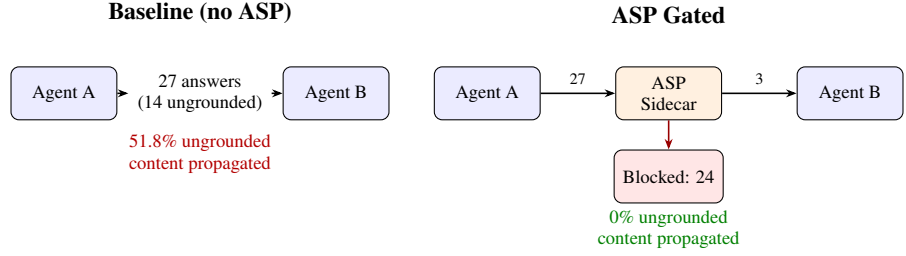

\subsection{Pipeline Results}
Table~\ref{tab:pipeline-results} and Figure~\ref{fig:pipeline-compare} summarize the pipeline benchmark on SmolLM3 (the model with the mixed failure profile, 44.4\% ungrounded).

\begin{table}[t]
\centering
\caption{Pipeline benchmark results (SmolLM3, 27 questions, risk assessment task)}
\label{tab:pipeline-results}
\begin{tabular}{lcc}
\toprule
\textbf{Metric} & \textbf{Baseline} & \textbf{ASP gated} \\
\midrule
Agent~A outputs reaching Agent~B & 27/27 (100\%) & 3/27 (11.1\%) \\
Ungrounded outputs reaching Agent~B & 14/27 (51.8\%) & 0/27 (0\%) \\
Blocked by sidecar & 0 & 24/27 (88.9\%) \\
Escalated to human review & 0 & 24 \\
\bottomrule
\end{tabular}
\end{table}

In the baseline condition, all 27 of Agent~A's outputs (including the 14 that are ungrounded) pass directly to Agent~B. Agent~B then makes risk assessments based on fabricated or poorly grounded upstream content, with no indication that the input was unreliable. In the ASP-gated condition, the sidecar blocks 24 of 27 outputs (including all 14 ungrounded ones), preventing Agent~B from reasoning on unreliable material. The 3 outputs that pass through are the least-drifted grounded answers.

\paragraph{Scope: SmolLM3 only}
We restrict the pipeline benchmark to SmolLM3 because Qwen and Dobby produce 0\% ungrounded outputs in the standalone benchmark (Section~\ref{sec:controller-results}); applying the sidecar to those agents would trivially pass 100\% of outputs through and would not exercise the gating decision. SmolLM3 is the only condition where the decision is non-trivial: 14 of 27 outputs are genuinely ungrounded, giving the sidecar a meaningful population to either block or pass. The relevant headline claim is that 100\% of \emph{ungrounded} outputs are blocked; the 88.9\% overall block rate is a function of SmolLM3's failure profile, not of the sidecar policy.

\paragraph{Effect of header visibility}
In the ASP-visible condition (where Agent~B receives the header but is not gated), Agent~B's judgments shift. When the header shows high grounding and low stochasticity, Agent~B produces specific risk assessments citing terms from Agent~A's answer. When the header shows low grounding or high stochasticity, Agent~B hedges or qualifies its judgment (e.g., ``MODERATE RISK'' with caveats rather than a confident assertion). This suggests that even without hard gating, making upstream quality visible to the downstream agent changes its behavior.

\paragraph{Implication for governance}
The pipeline benchmark demonstrates the sidecar model in practice. Without ASP, 51.8\% of upstream outputs would have propagated ungrounded content to a downstream decision agent. With the ASP sidecar, zero ungrounded outputs reach the downstream agent. The sidecar does not make SmolLM3 more capable (Agent~A's pass rate remains 0\%), but it prevents unreliable outputs from contaminating downstream decisions. This is the multi-agent version of the same principle demonstrated in the standalone benchmark: ASP changes how failure propagates, not how often models succeed.

\section{Operational Interpretation}

The benchmark supports two operational claims. First, ASP produces meaningful aggregate gains: 12/81 to 21/81 passes, with 10 recoveries. Second, the stronger claim is qualitative: ASP changes controller behavior in a way that matches the model's failure regime. Qwen and Dobby benefit from repair interventions, while SmolLM3 requires per-question triage between repair and containment.

\paragraph{Failure legibility}
This distinction is useful for XAI and trustworthy AI settings. A human supervisor does not only need a final answer; they need a system that can make its own failure state legible. Qwen's and Dobby's outputs are worth repairing, while SmolLM3's require per-question triage. A controller that can separate those cases is more useful than one that treats all non-passing drafts as equivalent. Without ASP, a supervisor reviewing a batch would see the raw results and have no structured way to prioritize which failures to investigate first. With ASP, they can immediately filter: 11 clean passes, 58 repair attempts (10 of which recovered to passing), and 12 containment events (none of which were worth additional effort).

\paragraph{Simple signals, narrow scope}
The signal formulas in Equations~\ref{eq:grounding}--\ref{eq:certainty} are intentionally simple weighted combinations rather than learned classifiers. This is a design choice: the signals must be auditable and reproducible without access to a calibration dataset. A consequence is that the controller cannot distinguish subtle failure modes (e.g., a plausible but wrong legal interpretation from a clearly nonsensical one) unless those modes also diverge in their token-overlap and citation profiles. For the current benchmark, this distinction holds cleanly: ungrounded answers have near-zero context overlap, while repairable answers have high overlap but missing terms. Whether this separation holds for more nuanced tasks (e.g., distinguishing correct legal reasoning from plausible-sounding but incorrect reasoning) remains an open question.

\paragraph{Drift as a complementary signal}
The drift monitor (Equation~\ref{eq:drift}) adds value by measuring how far the semantic vector deviates from an ideal QA distribution. On Qwen, drift stays below 0.18 for most repair cases, reflecting answers that are close to the ideal profile but incomplete. On Dobby, drift is moderate for grounded answers and higher for partial failures, enabling the controller to distinguish repairable drafts from those approaching containment territory. On SmolLM3, drift spans both regimes within the same run (from moderate values on grounded questions to $d > 0.60$ on uncited paraphrases), enabling per-question triage. This range illustrates why a single global threshold would be insufficient: the controller needs to make per-answer decisions based on where each draft falls in the signal space.

\paragraph{Interpreting the coverage gains}
The benchmark also sharpens the meaning of ``improvement.'' On Qwen, term coverage rises from 36.7\% to 65.4\%, a 29-point gain that reflects substantially more complete answers. On Dobby, coverage rises from 48.8\% to 54.3\% alongside 4 new passes. These gains span both the intermediate quality metrics and end-task success. A legal reviewer who receives an answer covering 65\% of required terms is better positioned to fill in the gaps than one who receives an answer covering 37\% of required terms. When the answer also passes the strict automated rubric (as 21 of 81 now do), the reviewer's burden is further reduced.

\section{Limitations and Future Work}
\label{sec:future-work}

This study is intentionally narrow. 
\begin{itemize}
    \item the benchmark covers one agreement document and only 27 questions, so the results should be read as a targeted controller study rather than a broad measure of legal reasoning. 
    \item The pass rubric is strict but synthetic: it depends on citation match and required-term coverage rather than human judgment of answer usefulness. 
    \item The controller confusion matrices are benchmark-relative and not independent policy-accuracy measurements.
    \item SmolLM3 was evaluated on a different backend from Qwen and Dobby, which weakens strong cross-model comparisons. 
    \item The benchmark does not yet measure auditor performance directly, such as fault-localization speed or inter-rater agreement with and without ASP telemetry. 
    \item The signal computation (Equations~\ref{eq:grounding}--\ref{eq:stochasticity}) uses fixed hand-tuned weights rather than learned coefficients, so the thresholds may not generalize beyond this document family.
    \item The adaptive assumption decay rate (Equation~\ref{eq:decay-rate}) and exponential weight dampening (Equation~\ref{eq:assumption-weight}) are designed for multi-turn conversations where assumptions accumulate over many exchanges; the current single-question benchmark (at most 3 drafts per question) does not exercise them over long enough trajectories for decay to influence routing. Similarly, the stochasticity-modulated centroid (Equation~\ref{eq:centroid}) has limited impact when the trajectory is only 1--3 steps long. \textbf{Three ablations are needed to validate these mechanisms} 
    \begin{itemize}
        \item Comparing adaptive $\lambda_t$ against a fixed decay rate
        \item Comparing centroid-based drift against fixed-ideal-only drift
        \item Demonstrating a trajectory where assumption decay below the 0.18 floor changes a routing decision. 
    \end{itemize}
    These ablations require longer multi-turn benchmarks and are planned for future work.
    \item The ideal QA reference vector $\mathbf{v}^*$ was hand-picked rather than fit to a held-out set of passing answers. Together with the routing thresholds $\tau_{\text{warn}}$ and $\tau_{\text{crit}}$, it determines essentially every routing decision on Dobby and most of them on SmolLM3. 

Finally, the Argent Framework \citep{sharma_argent} outlines a learned latent-channel variant of ASP in which encoders project the protocol semantics into disentangled channels consumed via cross-attention rather than text re-parsing. This paper evaluates only the surfaced text-level controller; validating the latent variant remains future work.
\end{itemize}
\section{Conclusion}

The results support three claims. 

\begin{itemize}
    \item ASP produces meaningful standalone gains, raising aggregate passes from 12/81 to 21/81 with 10 fail-to-pass recoveries.
    \item The same controller adapts its operational role to each model: repair on Qwen and Dobby, per-question triage on SmolLM3.
    \item In multi-agent mode the ASP sidecar prevents 100\% of ungrounded upstream outputs from reaching a downstream decision agent, demonstrating that the protocol mitigates semantic drift across agent boundaries.
\end{itemize}

\bibliographystyle{unsrtnat}
\bibliography{references}

\end{document}